\documentclass[10pt,twocolumn,letterpaper]{article}

\usepackage{iccv}
\usepackage{times}
\usepackage{epsfig}
\usepackage{graphicx}
\usepackage{amsmath}
\usepackage{amssymb}
\usepackage{multirow}
\usepackage{diagbox}
\usepackage{amsmath,booktabs}

\usepackage[pagebackref=true,breaklinks=true,letterpaper=true,colorlinks,bookmarks=false]{hyperref}

\iccvfinalcopy 

\def\iccvPaperID{2799} 

\ificcvfinal\fi
\begin{document}

\title{PU-GAN: a Point Cloud Upsampling Adversarial Network}

\author{Ruihui Li$^{1}$\quad Xianzhi Li$^{1,3}$\quad Chi-Wing Fu$^{1,3}$\quad Daniel Cohen-Or$^2$\quad Pheng-Ann Heng$^{1,3}$\\
	\hspace{-10mm}$^1$The Chinese University of Hong Kong \hspace{10mm} $^2$ Tel Aviv University\\
	$^3$Guangdong Provincial Key Laboratory of Computer Vision and Virtual Reality Technology,\\
	Shenzhen Institutes of Advanced Technology, Chinese Academy of Sciences, China\\
	\hspace{-10mm}{\tt\small \{lirh,xzli,cwfu,pheng\}@cse.cuhk.edu.hk}\hspace{10mm}{\tt\small dcor@mail.tau.ac.il}\qquad
}
\newcommand{\TODO}[1]{{\color{red}{[TODO: #1]}}}
\newcommand{\phil}[1]{{\color[rgb]{0.3,0.7,0.3}{[PH: #1]}}}
\newcommand{\rh}[1]{#1}
\newcommand{\xz}[1]{{\color[rgb]{1.0,0.6,0}{[XZ: #1]}}}
\newcommand{\dc}[1]{{\color{red}{[DC: #1]}}}
\newcommand{\para}[1]{\vspace{.05in}\noindent\textbf{#1}}
\def\ie{\emph{i.e.}}
\def\eg{\emph{e.g.}}
\def\etal{{\em et al.}}
\def\etc{{\em etc.}}

\maketitle

\begin{abstract}
Point clouds acquired from range scans are often sparse, noisy, and non-uniform.
%
This paper presents a new point cloud upsampling network called PU-GAN \footnote[1]{\href{https://liruihui.github.io/publication/PU-GAN/}{https://liruihui.github.io/publication/PU-GAN/}}, which is formulated based on a generative adversarial network (GAN), to learn a rich variety of point distributions from the latent space and upsample points over patches on object surfaces.
%
To realize a working GAN network, we construct an up-down-up expansion unit in the generator for upsampling point features with error feedback and self-correction, and formulate a self-attention unit to enhance the feature integration.
Further, we design a compound loss with adversarial, uniform and reconstruction terms, to encourage the discriminator to learn more latent patterns and enhance the output point distribution uniformity.
%
Qualitative and quantitative evaluations demonstrate the quality of our results over the state-of-the-arts in terms of distribution uniformity, proximity-to-surface, and 3D reconstruction quality.
%
%
%
%
%
\end{abstract}


\section{Introduction}
\label{sec:intro}

\begin{figure}[!t]
	\centering
	\includegraphics[width=0.95\linewidth]{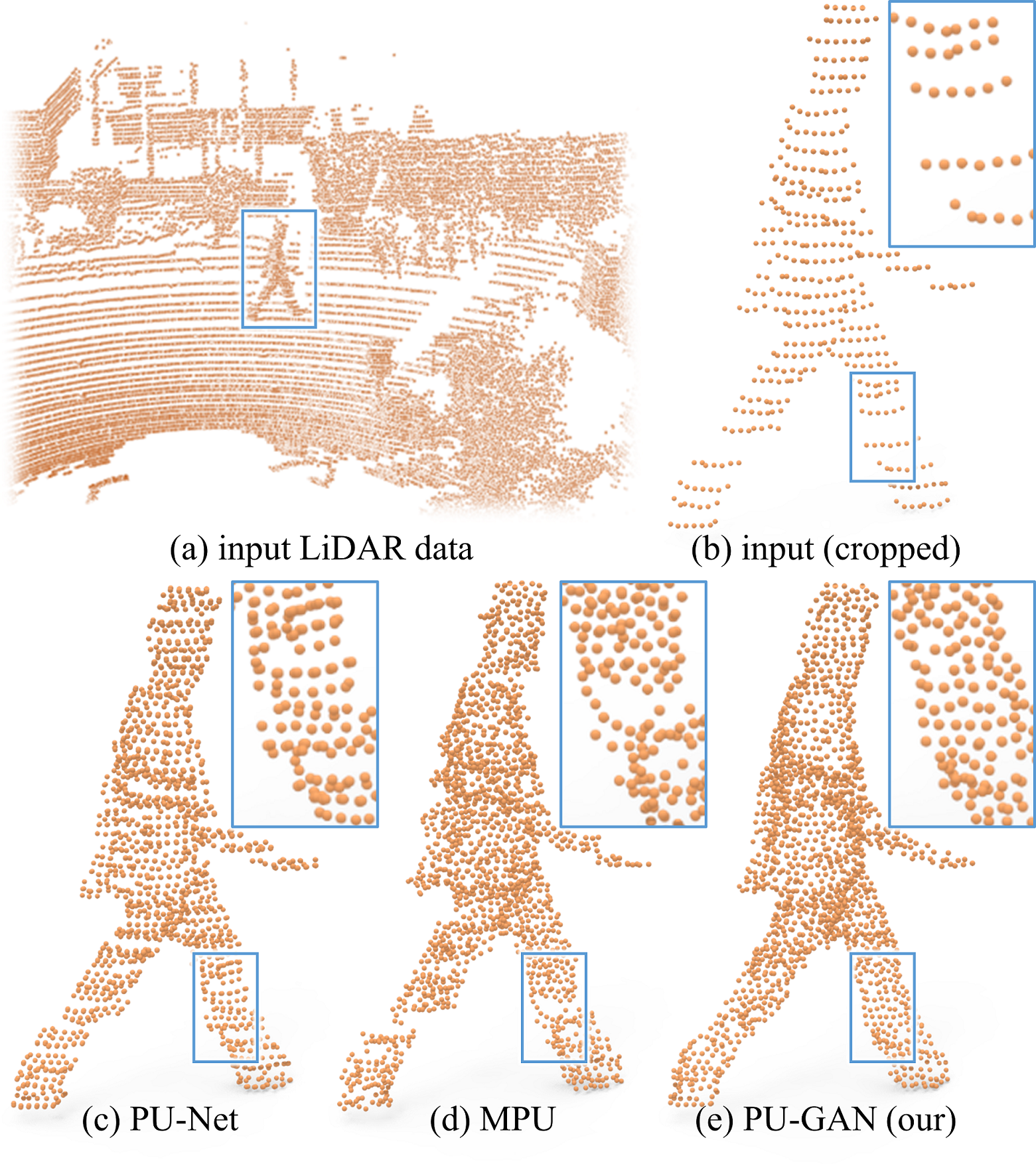}
	\caption{Upsampling (b) a point set cropped from (a) the real-scanned KITTI dataset~\cite{geiger2013vision} by using (c) PU-Net~\cite{yu2018pu} (CVPR 2018), (d) MPU~\cite{yifan2018patch} (CVPR 2019), and (e) our PU-GAN.
Note the distribution non-uniformity and sparsity in the input.}
	\label{fig:teaser}
	\vspace*{-2mm}
\end{figure}

Point clouds are the standard outputs from 3D scanning.
In recent years, they are gaining more and more popularity as a compact representation for 3D data, and as an effective means for processing 3D geometry.
However, raw point clouds produced from depth cameras and LiDAR sensors are often sparse, noisy, and non-uniform.
This is evidenced in various public benchmark datasets, such as KITTI~\cite{geiger2013vision}, SUN RGB-D~\cite{song2015sun}, and ScanNet~\cite{dai2017scannet}.
Clearly, the raw data is required to be amended, before it can be effectively used for rendering, analysis, or general processing.


%


Given a sparse, noisy, and non-uniform point cloud, our goal is to upsample it and generate a point cloud that is \emph{dense}, \emph{complete}, and \emph{uniform}, as a faithful representation of the underlying surface.
%
These goals are very challenging to achieve, given such an imperfect input, since, apart from upsampling the data, we need to fill small holes and gaps in data, while improving the point distribution uniformity.
%

Early methods~\cite{alexa2003computing,lipman2007parameterization,huang2009consolidation,huang2013edge,wu2015deep} for the problem
are optimization-based, where various shape priors are used to constrain the point cloud generation.
Recently, deep neural networks brought the promise of data-driven approaches to the problem.
Network architectures, including PU-Net~\cite{yu2018pu}, EC-Net~\cite{yu2018ec}, and very recently, MPU~\cite{yifan2018patch}, have demonstrated the advantage of upsampling point clouds through learning.
%
%
However, these networks may not produce plausible results from low-quality inputs that are particularly sparse and non-uniform;
in Figure~\ref{fig:teaser} (c)-(e), we show a typical example that demonstrates the advantage of our method, which unlike the other networks, it combines completion and uniformity together with upsampling.

In this work, we present a new point cloud upsampling framework, namely PU-GAN, that combines upsampling with data amendment capability.
The key contribution is an adversarial network to enable us to train a generator network to learn to produce a rich variety of point distributions from the latent space, and also a discriminator network to help implicitly evaluate the point sets produced from the generator against the latent point distribution.
Particularly, the adversarial learning strategy of the GAN framework can regularize the predictions from a global perspective and implicitly penalize outputs that deviate from the target.



However, successfully training a working GAN framework is known to be challenging~\cite{goodfellow2014distinguishability,radford2015unsupervised}, particularly the difficulty to balance between the generator and discriminator and to avoid the tendency of poor convergence.
%
Thus, we first improve the point generation quality, or equivalently the feature expansion capability, of the generator, by constructing an up-down-up unit to expand point features by upsampling also their differences for self-correction.
Besides, we formulate a self-attention unit to enhance the feature integration quality.
Lastly, we further design a compound loss to train the network end-to-end with adversarial, uniform, and reconstruction terms to enhance the distribution uniformity of the results and encourage the discriminator to learn more latent patterns in the target distribution.

To evaluate the upsampling quality of PU-GAN, we employ four metrics to assess its performance against the state-of-the-arts on a variety of synthetic and real-scanned data.
Extensive experimental results show that our method outperforms others in terms of distribution uniformity, proximity-to-surface, and 3D reconstruction quality.

\section{Related Work}
\label{sec:bg}


\para{Optimization-based upsampling.} \
To upsample a point set, a pioneering method by Alexa~\etal~\cite{alexa2003computing} inserts points at Voronoi diagram vertices in the local tangent space.
Later, Lipman~\etal~\cite{lipman2007parameterization} introduced the locally optimal projection (LOP) operator to resample points and reconstruct surfaces based on an $L_1$ norm.
Soon after, Huang~\etal~\cite{huang2009consolidation} devised a weighted LOP with an iterative normal estimation to consolidate point sets with noise, outliers, and non-uniformities.
Huang~\etal~\cite{huang2013edge} further developed a progressive method called EAR for edge-aware resampling of point sets.
Later, Wu~\etal~\cite{wu2015deep} proposed a consolidation method to fill large holes and complete missing regions by introducing a new deep point representation.
Overall, these methods are {\em not\/} data-driven; they heavily rely on priors,~\eg, the assumption of smooth surface, normal estimation,~\etc

\begin{figure*}[t]
	\centering
	\includegraphics[width=0.88\linewidth]{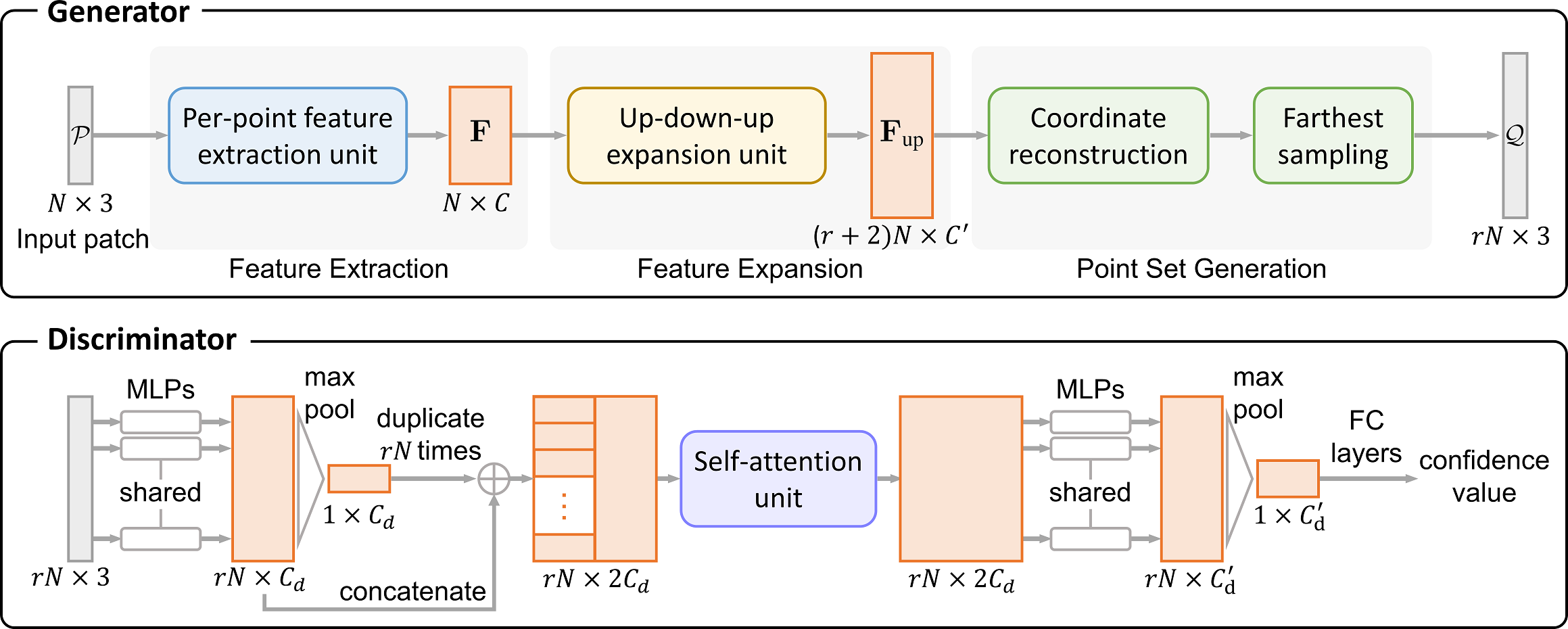}
	\caption{Overview of PU-GAN's generator and discriminator architecture.
Note that $N$ is the number of points in input $\mathcal{P}$; $r$ is the upsampling rate; and $C$, $C'$, $C_d$, and $C_d'$ are the number of feature channels that are 480, 128, 64, and 256, respectively, in our implementation.}
	\label{fig:framework}
	\vspace*{-2mm}
\end{figure*}


\para{Deep-learning-based upsampling.} \
In recent few years, several deep neural networks have been designed to learn features directly from point sets,~\eg, PointNet~\cite{qi2016pointnet}, PointNet++~\cite{qi2017pointnet++}, SpiderCNN~\cite{xu2018spidercnn}, KCNet~\cite{shen2018mining}, SPLATNet~\cite{su2018splatnet}, PointCNN~\cite{li2018pointcnn}, PointConv~\cite{hua2017pointwise}, PointWeb~\cite{zhao2019pointweb}, DGCNN~\cite{wang2018dynamic}, \emph{etc}.
\rh{Different from point completion ~\cite{yuan2018pcn,gurumurthy2019high} which
generates the entire object from partial input, point cloud upsampling tends to improve the point distribution uniformity within local patches.}
Based on the PointNet++ architecture, Yu~\etal~\cite{yu2018pu} introduced a deep neural network PU-Net to upsample point sets.
PU-Net works on patches and expands a point set by mixing-and-blending point features in the feature space.
Later, they introduced EC-Net~\cite{yu2018ec} for edge-aware point cloud upsampling by formulating an edge-aware joint loss function to minimize the point-to-edge distances.
%
Very recently, Wang~\etal~\cite{yifan2018patch} presented a multi-step progressive upsampling (MPU) network to further suppress noise and retain details.
%
In this work, we present a new method to upsample point clouds by formulating a GAN framework, enabling us to generate higher-quality point samples with completion and uniformity.




\para{GAN-based 3D shape processing.} \
Compared with the conventional CNN, the GAN framework makes use of a discriminator to implicitly learn to evaluate the point sets produced from the generator.
%
Such flexibility particularly benefits generative tasks~\cite{ledig2017photo,isola2017image,zhang2017stackgan}, since we hope the generator can produce a rich variety of output patterns.
%

Recently, there are some inspiring works in applying GAN to 3D shape processing.
Most of them focus on generating 3D objects either from a probabilistic space~\cite{wu2016learning} or from 2D images~\cite{yang20173d,gwak2017weakly,smith2017improved}.
Moreover, Wang~\etal~\cite{wang2017shape} introduced a 3D-ED-GAN for shape completion given a corrupted 3D scan as input.
However, these methods can only consume 3D volumes as inputs or output shapes in voxel representations.
Achlioptas~\etal~\cite{achlioptas2018learning} first adapted the GAN model to operate on raw point clouds for enhancing the representation learning.
%
To the best of our knowledge, no prior works develop GAN models for point cloud upsampling.

\section{Method}
\label{sec:method}

\subsection{Overview}
\label{subsec:overview}
Given an unordered sparse point set $\mathcal{P}=\{p_i\}_{i=1}^N$ of $N$ points, we aim to generate a dense point set $\mathcal{Q}=\{q_i\}_{i=1}^{rN}$ of $rN$ points, where $r$ is the upsampling rate.
While output $\mathcal{Q}$ is not necessarily a superset of $\mathcal{P}$, we want it to satisfy two requirements:
(i) $\mathcal{Q}$ should describe the {\em same underlying geometry\/} of a latent target object as $\mathcal{P}$, so points in $\mathcal{Q}$ should lie on and cover the target object surface; and
(ii) points in $\mathcal{Q}$ should be {\em uniformly-distributed\/} on the target object surface, even for sparse and non-uniform input $\mathcal{P}$.

%
Figure~\ref{fig:framework} shows the overall network architecture of PU-GAN, where the generator produces dense output $\mathcal{Q}$ from sparse input $\mathcal{P}$,
and the discriminator aims to find the fake generated ones.
Following, we first elaborate on the architecture of the generator and discriminator (Section~\ref{subsec:network}).
We then present two building blocks in our architecture: the up-down-up expansion unit (Sections~\ref{subsec:projection}) and the self-attention unit (Section~\ref{subsec:attention}).
Lastly, we present the patch-based training strategy with the compound loss (Section~\ref{subsec:training}).


\subsection{Network Architecture}
\label{subsec:network}

\subsubsection{Generator}
As shown on top of Figure~\ref{fig:framework}, our generator network has three components to successively process input $\mathcal{P}$:

\para{The feature extraction component} aims to extract features $\mathbf{F}$ from input $\mathcal{P}$ of $N \times d$, where $d$ is the number of dimensions in the input point attributes,~\ie, coordinates, color, normal,~\etc\
Here, we focus on the simplest case with $d=3$, considering only the 3D coordinates, and adopt the recent feature extraction method in~\cite{yifan2018patch}, where dense connections are introduced to integrate features across different layers.
%

\para{The feature expansion component} expands $\mathbf{F}$ to produce the expanded features $\mathbf{F_{\text{up}}}$; here, we design the \emph{up-down-up expansion unit} (see Figure~\ref{fig:framework} (top)) to enhance the feature variations in  $\mathbf{F_{\text{up}}}$, enabling the generator to produce more diverse point distributions; see Section~\ref{subsec:projection} for details.

\para{The point set generation component} first regresses a set of 3D coordinates from $\mathbf{F_{\text{up}}}$ via a set of multilayer perceptrons (MLPs).
Since the feature expansion process is still local, meaning that the features in $\mathbf{F_{\text{up}}}$ (or equivalently, points in the latent space) are intrinsically close to the inputs, we thus include a farthest sampling step in the network to retain only $rN$ points that are further away from one another; see the green boxes in Figure~\ref{fig:framework}.
To allow this selection, when we expand $\mathbf{F}$ to $\mathbf{F_{\text{up}}}$, we actually generate $(r+2)N$ features in  $\mathbf{F_{\text{up}}}$.
This strategy further enhances the point set distribution uniformity, particularly from a global perspective.
%

\subsubsection{Discriminator}
The goal of the discriminator is to distinguish whether its input (a set of $rN$ points) is produced by the generator.
To do so, we first adopt the basic network architecture in~\cite{yuan2018pcn} to extract the global features, since it efficiently combines the local and global information, and ensures a lightweight network.
To improve the feature learning, we add a self-attention unit (see Section~\ref{subsec:attention}) after the feature concatenation; see the middle part in Figure~\ref{fig:framework} (bottom).
Compared with the basic MLPs, this attention unit helps enhance the feature integration and improve the subsequent feature extraction capability.
%
Next, we apply a set of MLPs and a max pooling to produce the global features, and further regress the final confidence value via a set of fully connected layers.
If this confidence value is close to 1, the discriminator predicts that the input likely comes from a target distribution with high confidence, and otherwise from the generator.


\subsection{Up-down-up Expansion Unit}
\label{subsec:projection}
To upsample a point set, PU-Net~\cite{yu2018pu} duplicates the point features and uses separate MLPs to process each copy independently.
However, the expanded features would be too similar to the inputs, so affecting the upsampling quality.
%
Rather than a single-step expansion, the very recent MPU method~\cite{yifan2018patch} breaks a $16$$\times$ upsampling network into four successive $2$$\times$ upsampling subnets to progressively upsample in multiple steps.
Though details are better preserved in the upsampled results, the training process is complex and requires more subsets for a higher upsampling rate.

\begin{figure}[!t]
	\centering
	\includegraphics[width=0.98\linewidth]{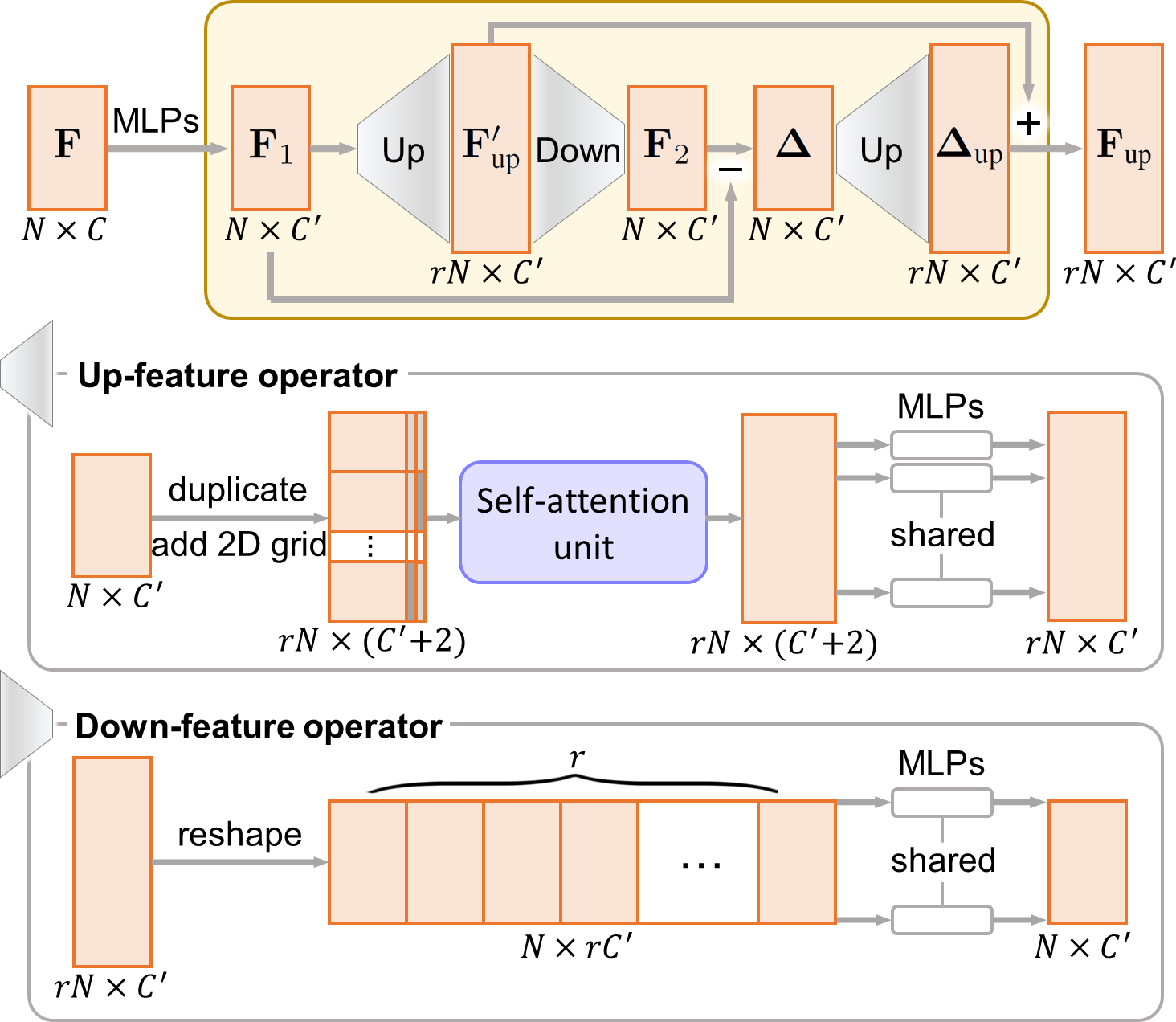}
	\caption{The up-down-up expansion unit (top), up-feature operator (middle), and down-feature operator (bottom).}
	\label{fig:block}
	\vspace*{-2mm}
\end{figure}

In this work, we construct an up-down-up expansion unit to expand the point features.
To do so, we first upsample the point features (after MLPs) to generate $\mathbf{F}_{\text{up}}^\prime$ and downsample it back (see Figure~\ref{fig:block} (top)); then, we compute the difference (denoted as $\Delta$) between the features before the upsampling and after the downsampling.
By also upsampling $\Delta$ to $\Delta_{\text{up}}$, we add $\Delta_{\text{up}}$ to $\mathbf{F}_{\text{up}}^\prime$ to self-correct the expanded features;
see again Figure~\ref{fig:block} (top) for the whole procedure.
Such a strategy not only avoids tedious multi-step training, but also facilitates the production of fine-grained features.

Next, we discuss the up-feature and down-feature operators in the up-down-up expansion unit (see also Figure~\ref{fig:block}):

\para{Up-feature operator.} \
To upsample the point features $r$ times, we should increase the variations among the duplicated features.
This is equivalent to pushing the new points away from the input ones.
After we duplicate the input feature map (of $N$ feature vectors and $C'$ channels) $r$ times, we adopt the 2D grid mechanism in FoldingNet~\cite{yang2018foldingnet} to generate a unique 2D vector per feature-map copy, and append such vector to each point feature vector in the same feature-map copy; see Figure~\ref{fig:block} (middle).
Further, we use the self-attention unit (see Section~\ref{subsec:attention}) followed by a set of MLPs to produce the output upsampled features.

\para{Down-feature operator.} \
To downsample the expanded features, we reshape the features and then use a set of MLPs to regress the original features; see Figure~\ref{fig:block} (bottom).

\subsection{Self-attention Unit}
\label{subsec:attention}

\begin{figure}[!t]
	\centering
	\includegraphics[width=0.75\linewidth]{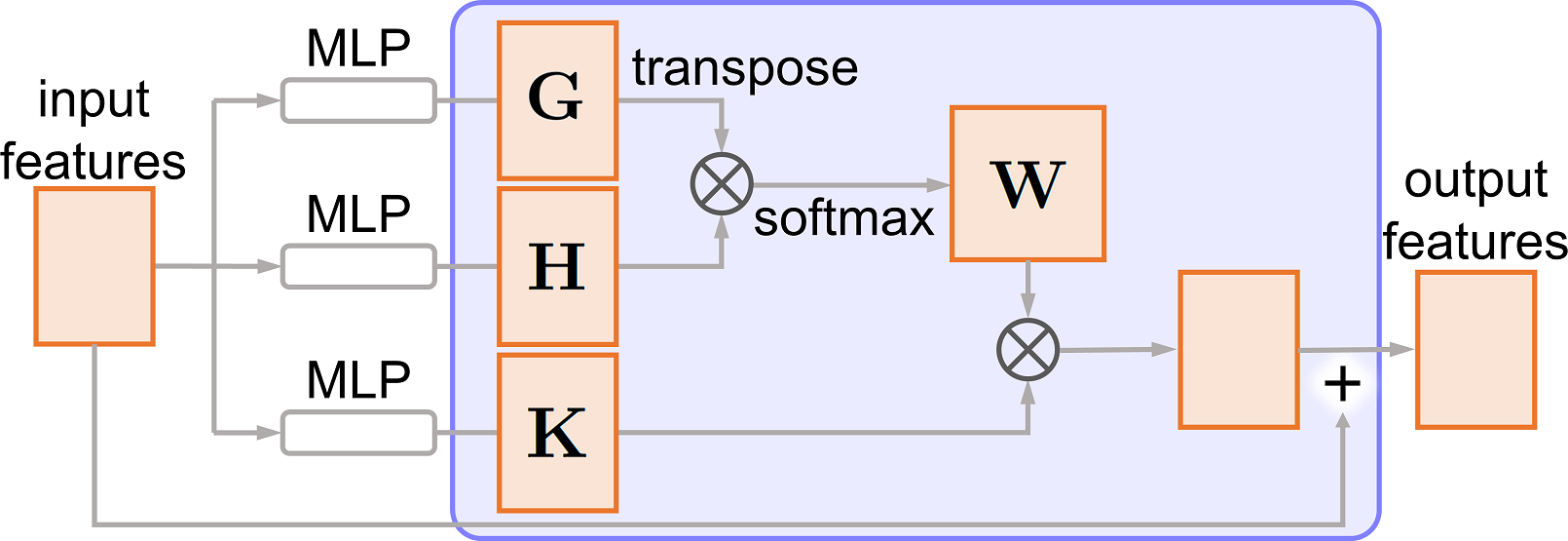}
	\caption{Illustration of the self-attention unit.}
	\label{fig:attention}
	\vspace*{-2mm}
\end{figure}

\rh{To introduce long-range context dependencies to enhance feature integration after concatenation}, we adopt the self-attention unit~\cite{zhang2019self} in the generator (see Figure~\ref{fig:block} (middle)), as well as in the discriminator (see Figure~\ref{fig:framework} (bottom)).
%
Figure~\ref{fig:attention} presents its architecture.
Specifically, we transform the input features into $\mathbf{G}$ and $\mathbf{H}$ via two separate MLPs, and then generate the attention weights $\mathbf{W}$ from $\mathbf{G}$ and $\mathbf{H}$ by
\begin{equation}
\mathbf{W} = f_{\text{softmax}}(\mathbf{G}^T \mathbf{H}) \ ,
\end{equation}
where $f_{\text{softmax}}$ denotes the softmax function.
After that, we obtain the weighted features $\mathbf{W}^T\mathbf{K}$, where $\mathbf{K}$ is the set of features extracted from the input via another MLP.
Lastly, we generate the output features, which is the sum of the input features and the weighted features.

%
%



\subsection{Patch-based End-to-end Training}
\label{subsec:training}

\subsubsection{Training data preparation}
\label{subsubsec:train_data}




\begin{figure}
	\centering
	\includegraphics[width=0.9\linewidth]{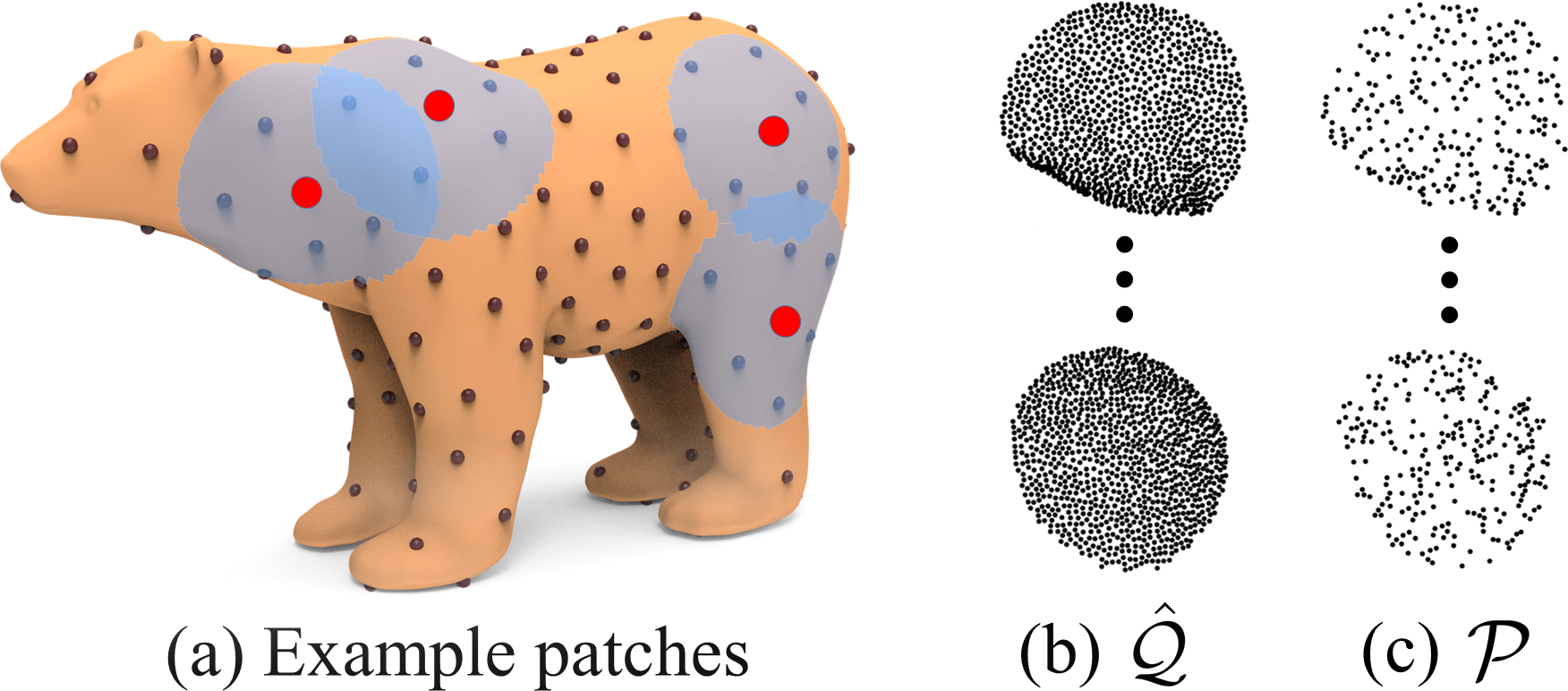}
	\caption{(a) Seed points (black dots) and patches (blue disks) on a 3D mesh in training data.
(b) \& (c) Example $\mathcal{\hat{Q}}$ and $\mathcal{P}$ on patches.}
	\label{fig:patch_segment}
	\vspace*{-2mm}
\end{figure}

We trained our network to upsample local groups of points over patches on object surface.
%
Specifically, for each 3D mesh (normalized in a unit sphere) in training set (see Section~\ref{subsec:dataset}), we use randomly
find 200 seed positions on each mesh surface, geodesically grow a patch from each seed (each patch occupies $\sim$5\% of the object surface), and then normalize each patch within a unit sphere; see Figure~\ref{fig:patch_segment}(a).
On each patch, we further use Poisson disk sampling~\cite{corsini2012efficient} to generate $\mathcal{\hat{Q}}$, which is a target point set of $rN$ points on the patch.
During the training, we generate the network input $\mathcal{P}$ by randomly selecting $N$ points on-the-fly from $\mathcal{\hat{Q}}$.

\subsubsection{Loss functions}
\label{subsubsec:loss}

We now present the compound loss designed for training PU-GAN in an end-to-end fashion.



\para{Adversarial loss.} \
To train the generator network $G$ and discriminator network $D$ in an adversarial manner, we use the least-squared loss~\cite{mao2017least} as our adversarial loss:
\begin{eqnarray}
\mathcal{L}_{\text{gan}}(G)&=&\frac{1}{2}[D(\mathcal{Q})-1]^2 \\
\text{and} \ \
\mathcal{L}_{\text{gan}}(D)&=&\frac{1}{2}[D(\mathcal{Q})^2 + (D(\mathcal{\hat{Q}})-1)^2],
\end{eqnarray}
where
$D(\mathcal{Q})$ is the confidence value predicted by $D$ from generator output $\mathcal{Q}$.
During the network training, $G$ aims to generate $\mathcal{Q}$ to fool $D$ by minimizing $\mathcal{L}_{\text{gan}}(G)$, while $D$ aims to minimize $\mathcal{L}_{\text{gan}}(D)$ to learn to distinguish $\mathcal{Q}$ from $\mathcal{\hat{Q}}$.
%


\para{Uniform loss.} \
The problem of learning to generate point sets in 3D is complex with an immense space for exploration during the network training.
Particularly, we aim for uniform distributions; using the adversarial loss alone is hard for the network to converge well.
Thus, we formulate a uniform loss to evaluate $\mathcal{Q}$ from the generator, aiming to improve the generative ability of the generator.
%

\begin{figure}[t]
	\centering
	\includegraphics[width=1.0\linewidth]{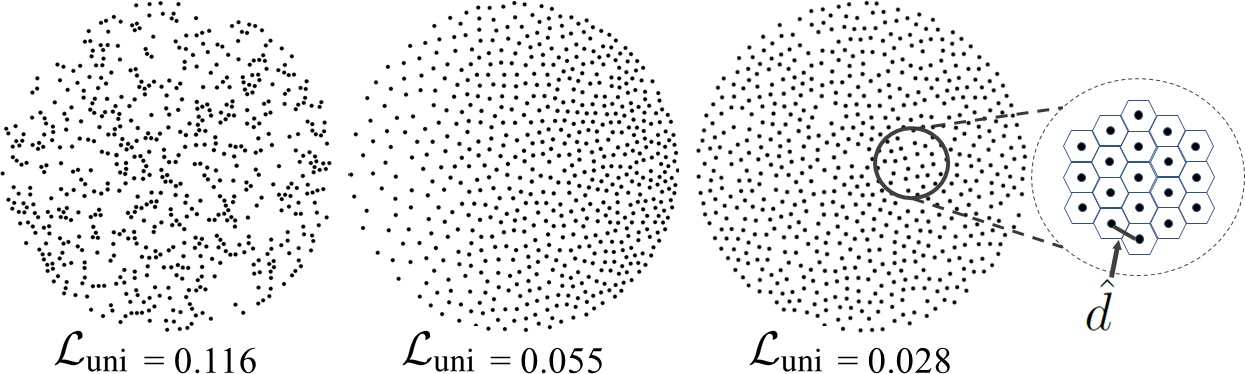}
	\caption{Example point sets with same number of points (625) but different point distribution patterns; $\mathcal{L_{\text{uni}}}$ is computed with $p$$=$$1\%$.}
	\label{fig:uniformity}
	\vspace*{-2mm}
\end{figure}

To evaluate a point set's uniformity, the NUC metric in PU-Net~\cite{yu2018pu} crops equal-sized disks on the object surface and counts the number of points in the disks.
So, the metric neglects the local clutter of points.
Figure~\ref{fig:uniformity} shows three patches of points with very different point distributions; since they contain the same number of points, the NUC metric cannot distinguish their distribution uniformity.

Our method evaluates $\mathcal{Q}$ (a patch of $rN$ points) during the network training by first using the farthest sampling to pick $M$ seed points in $\mathcal{Q}$ and using a ball query of radius $r_d$ to crop a point subset (denoted as $S_j$, $j=1..M$) in $\mathcal{Q}$ at each seed.
Here, we use a small $r_d$, so $S_j$ roughly lies on a small local disk of area $\pi r_d^2$ on the underlying surface.
On the other hand, since we form patches by geodesics and normalize them in a unit sphere, patch area is $\sim$$\pi 1^2$.
So, the expected percentage of points in $S_j$ (denoted as $p$) is $\pi r_d^2 / \pi 1^2 = r_d^2$, and the expected number of points in $S_j$ (denoted as $\hat{n}$) is $rNp$.
Hence, we follow the chi-squared model to measure the deviation of $|S_j|$ from $\hat{n}$, and define
\begin{equation}
U_{\text{imbalance}}(S_j) = \frac{(|S_j|-\hat{n})^2}{\hat{n}} \ .
\end{equation}

To account for local point clutter, for each point in $S_j$, we find its distance to the nearest neighbor, denoted as $d_{j,k}$ for the $k$-th point in $S_j$.
If $S_j$ has a uniform distribution, the expected point-to-neighbor distance $\hat{d}$ should be roughly $\sqrt{\frac{2\pi r_d^2}{|S_j| \sqrt{3}}}$, which is derived based on the assumption that $S_j$ is flat and neighboring points are hexagonal; see Figure~\ref{fig:uniformity} (right) for an illustration and supplemental material for the derivation.
Again, we follow the chi-squared model to measure the deviation of $d_{j,k}$ from $\hat{d}$, and define
\begin{equation}
U_{\text{clutter}}(S_j) = \sum_{k=1}^{|S_j|} \frac{(d_{j,k}-\hat{d})^2}{\hat{d}} \ .
\end{equation}

Here, $U_{\text{clutter}}$ accounts for the local distribution uniformity, while $U_{\text{imbalance}}$ accounts for the nonlocal uniformity to encourage better point coverage.
Putting them together, we formulate the uniform loss as
\vspace*{-1.5mm}
\begin{equation}
\label{eq:uniform}
\mathcal{L_{\text{uni}}} = \sum_{j=1}^{M}U_{\text{imbalance}}(S_j) \cdot U_{\text{clutter}}(S_j) \ .
\vspace*{-1mm}
\end{equation}
Figure~\ref{fig:uniformity} shows three example point sets with the same number of points but different point distribution patterns.
Using $\mathcal{L_{\text{uni}}}$, we can distinguish the point uniformity among them.

\para{Reconstruction loss.} \
Both the adversarial and uniform losses do not encourage the generated points to lie on the target surface.
Thus, we include a reconstruction loss using the Earth Mover's distance (EMD)~\cite{fan2016point}:
%
\vspace*{-1.5mm}
\begin{equation}
	\mathcal{L_{\text{rec}}} = \min_{\phi:\mathcal{Q}\rightarrow \mathcal{\hat{Q}}} \sum_{q_i\in \mathcal{Q}} \|q_i-\phi(q_i)\|_2,
	\vspace*{-1mm}
\end{equation}
where 
$\phi:\mathcal{Q}\rightarrow \mathcal{\hat{Q}}$ is the bijection mapping.


\para{Compound loss.} \
Overall, we train PU-GAN end-to-end by minimizing $\mathcal{L}_G$ for generator and  $\mathcal{L}_D$ for discriminator:
\begin{eqnarray}
\label{eq:compound}
	\mathcal{L}_G&=& \lambda_{\text{gan}}\mathcal{L_{\text{gan}}}(G) +\lambda_{\text{rec}}\mathcal{L_{\text{rec}}} + \lambda_{\text{uni}}\mathcal{L_{\text{uni}}}, \\
	\text{and} \ \
    \mathcal{L}_D&=&\mathcal{L}_{\text{gan}}(D),
\end{eqnarray}
where $\lambda_{\text{gan}}$, $\lambda_{\text{rec}}$, and $\lambda_{\text{uni}}$ are weights.
During the training process, $G$ and $D$ are optimized alternatively.


\section{Experiments}
\label{sec:experiment}


\begin{figure*}[t]
	\centering
	\includegraphics[width=0.99\linewidth]{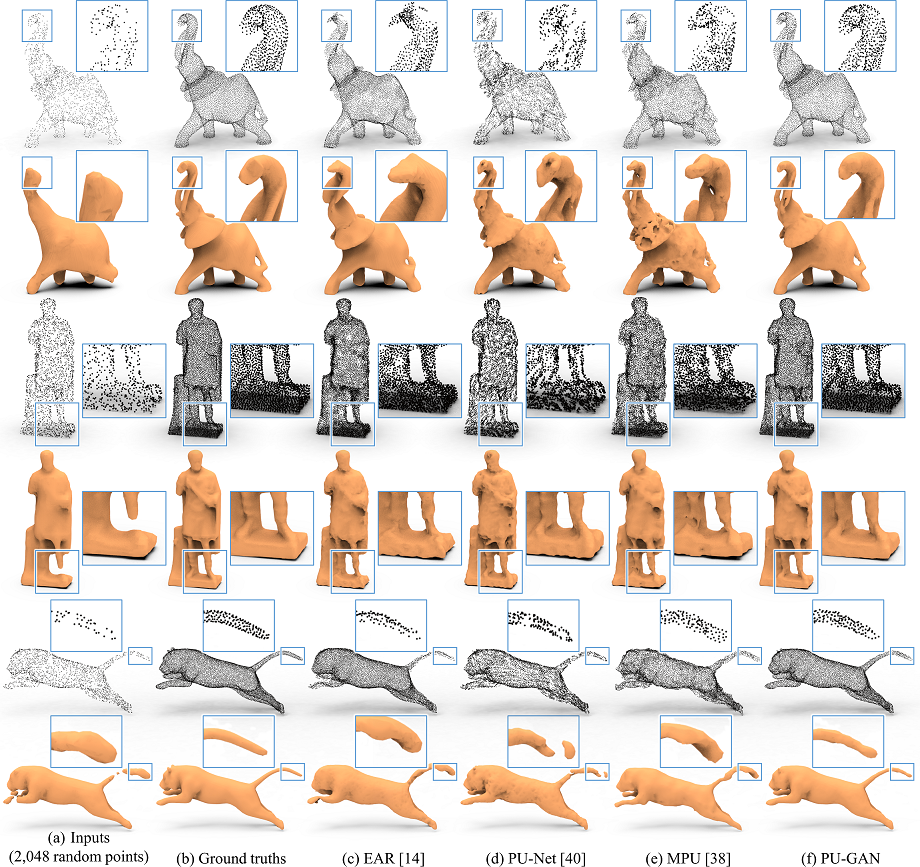}
	\caption{Comparing point set upsampling (x4) and surface reconstruction results produced with different methods (c-f) from inputs (a).}
	\label{fig:visualComparison}
	\vspace*{-2mm}
\end{figure*}


\begin{figure*}[t]
	\centering
	\includegraphics[width=1.0\linewidth]{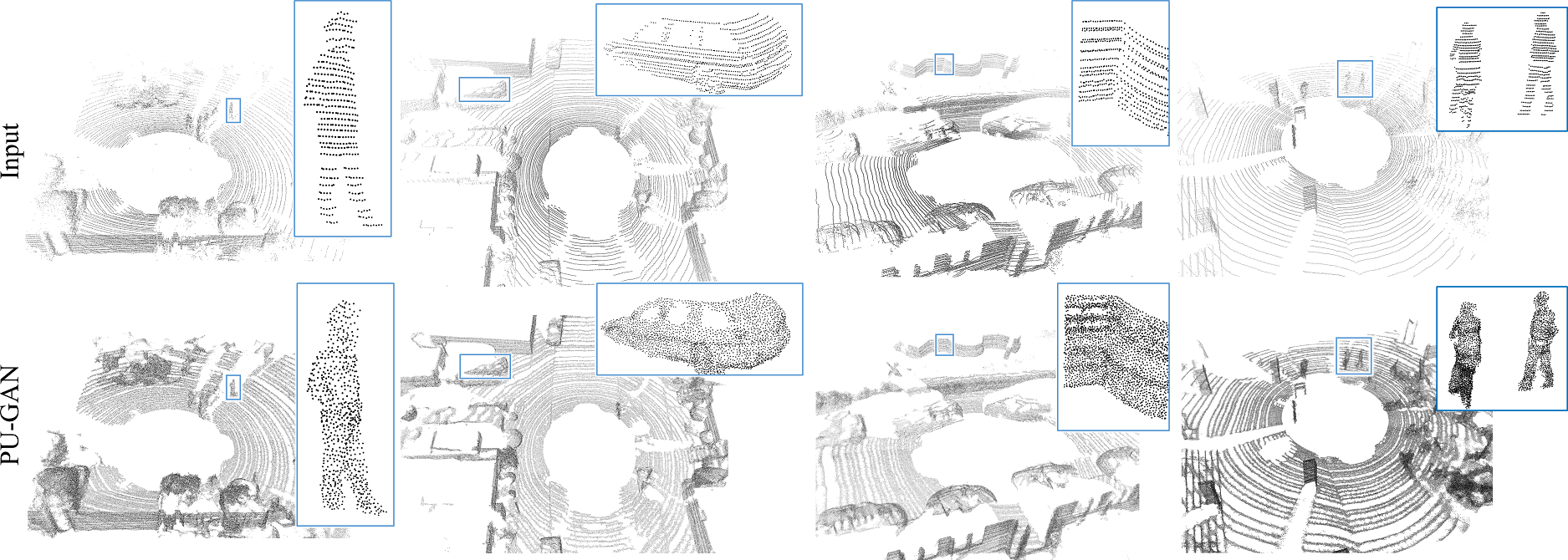}
	\caption{Using PU-GAN to upsample real-scanned point cloud data acquired by LiDAR.}
	\label{fig:realscan}
	\vspace*{-2mm}
\end{figure*}



\subsection{Datasets and Implementation Details}
\label{subsec:dataset}
We collected 147 3D models from the released datasets of PU-Net~\cite{yu2018pu} and MPU~\cite{yifan2018patch}, as well as from the Visionair repository~\cite{visionair}, covering a rich variety of objects, ranging from simple and smooth models (\eg, Icosahedron) to complex and high-detailed objects (\eg, Statue); see the supplemental material for all of them.
Among them, we randomly select 120 models for training and use the rest for testing.





In the training phase, we cropped 200 patches per training model (see Section~\ref{subsubsec:train_data}) and produced $24,000$ patches in total.
By default, we set $N=256$, $r=4$, and $M=50$.
Moreover, for the uniform loss, we cropped one set of $S_j$ with radius $r_d=\sqrt{p}$ for each $p \in$ \{ 0.4\%, 0.6\%, 0.8\%, 1.0\%, 1.2\% \}, compute Eq.~\eqref{eq:uniform} five times for each set, and then sum up the results as the $\mathcal{L_{\text{uni}}}$ term in Eq.~\eqref{eq:compound}.

To avoid overfitting in training, we augment the network inputs by random rotation, scaling, and point perturbation with Gaussian noise.
We trained the network for 100 epochs using the Adam algorithm~\cite{kingma2014adam} with a two time-scale update rule (TTUR)~\cite{heusel2017gans}.
We set the learning rates of generator and discriminator as 0.001 and 0.0001, respectively; after 50k iterations, we gradually reduce both rates by a decay rate of 0.7 per 50k iterations until $10^{-6}$.
The batch size is 28, and $\lambda_{\text{rec}}$, $\lambda_{\text{uni}}$, and $\lambda_{\text{gan}}$ are empirically set as 100, 10, and 0.5, respectively.
We implemented our network using TensorFlow and trained it on NVidia Titan Xp GPU.

During the testing, given a point set, we follow the patch-based strategies in MPU~\cite{yifan2018patch} and EC-Net~\cite{yu2018ec} to use the farthest sampling to pick seeds and extract a local patch of $N$ points per seed.
Then, we feed the patches to the generator and combine the upsampled results as the final output.


\subsection{Evaluation Metrics}
\label{subsec:evaluation}
We employ four evaluation metrics: (i) uniformity using $\mathcal{L_{\text{uni}}}$ in Eq.~\eqref{eq:uniform}, (ii) point-to-surface (P2F) distance using the testing models, (iii) Chamfer distance (CD), and (iv) Hausdorff distance (HD)~\cite{berger2013benchmark}.
For quantitative evaluation, we use Poisson disk sampling to sample 8,192 points as the ground truth (\eg, see Figure~\ref{fig:visualComparison}(b)) on each of the testing models, and randomly select 2,048 points as testing input.
To evaluate the uniformity of the test results (using Eq.~\eqref{eq:uniform}), we randomly pick $M=1,000$ seeds on each result, and instead of using ball query to crop $S_j$, we use the actual mesh (testing model) to geodesically find $S_j$ at each seed of different $p$ for higher-quality evaluation.
The lower the metric values are, the better the upsampling results are.


\subsection{Qualitative and Quantitative Comparisons}
\label{subsec:comparison}

We qualitatively and quantitatively compared PU-GAN with three state-of-the-art point set upsampling methods: EAR~\cite{huang2013edge}, PU-Net~\cite{yu2018pu}, and MPU~\cite{yifan2018patch}.
For EAR, we employed the released demo code and generated the best results by exhaustively fine-tuning every associated parameter.
For PU-Net and MPU, we used their public code and retrained their networks using our training data.

Table~\ref{tab:quanComparison} shows the quantitative comparison results.
Our PU-GAN achieves the lowest values consistently for all the evaluation metrics.
Particularly, the uniformity of our results stays the lowest for all different $p$, indicating that the points generated by PU-GAN are much more uniform than those produced by others over varying scales.

Besides quantitative results, we show point set upsampling and surface reconstruction (using~\cite{kazhdan2013screened}) results for various models in Figure~\ref{fig:visualComparison}.
Comparing the results produced by (f) our method and by (c-e) others, against (b) the ground truth points that are uniformly-sampled on the original testing models, we can see that other methods tend to produce more noisy and nonuniform point sets, thus leading to more artifacts in the reconstructed surfaces.
See, particularly, the blown-up views, showing that PU-GAN can produce more fine-grained details in the upsampled results,~\eg, elephant's nose (top) and tiger's tail (bottom).
More comparison results can be found in the supplemental material.



\subsection{Upsampling Real-scanned Data}

Figure~\ref{fig:realscan} shows upsampling results on LiDAR point clouds (downloaded from KITTI dataset~\cite{geiger2013vision}) produced by PU-GAN.
From the first row, we can see the sparsity and non-uniformity of the inputs.
Our PU-GAN can still fill some holes and output more uniform points in the results; please see the supplemental material for more results.

\begin{table}[!t]
\centering
\caption{Quantitative comparisons with the state-of-the-arts.}
\label{tab:quanComparison}
\resizebox{\linewidth}{!}{
	\begin{tabular}{@{\hspace{1mm}}c@{\hspace{1mm}}||@{\hspace{1mm}}c@{\hspace{3mm}}c@{\hspace{3mm}}c@{\hspace{3mm}}c@{\hspace{3mm}}c@{\hspace{1mm}}|@{\hspace{0.2mm}}c@{\hspace{1mm}}|@{\hspace{1mm}}c@{\hspace{1mm}}|@{\hspace{1mm}}c@{\hspace{1mm}}} \toprule[1pt]
	\multirow{2}*{Methods} & \multicolumn{5}{@{\hspace{1mm}}c@{\hspace{1mm}}|@{\hspace{0.2mm}}}{Uniformity for different $p$ ($10^{-3}$)} & P2F & CD & HD \\ \cline{2-6}
	& 0.4\% & 0.6\% & 0.8\% & 1.0\% & 1.2\% & ($10^{-3}$) & ($10^{-3}$) & ($10^{-3}$)\\ \hline \hline
	EAR~\cite{huang2013edge}  &  16.84  & 20.27  & 23.98  &  26.15  & 29.18   & 5.82 & 0.52  & 7.37   \\
	PU-Net~\cite{yu2018pu} &  29.74  & 31.33  & 33.86  & 36.94   & 40.43 & 6.84 & 0.72  & 8.94   \\
	MPU~\cite{yifan2018patch} & 7.51  & 7.41  & 8.35  & 9.62   & 11.13  & 3.96 & 0.49  & 6.11   \\ \hline
	PU-GAN & \textbf{3.38} & \textbf{3.49}  & \textbf{3.44}  & \textbf{3.91}  & \textbf{4.64} & \textbf{2.33} & \textbf{0.28}  & \textbf{4.64}    \\ \bottomrule[1pt]
	\end{tabular}}
\end{table}

\begin{table}[t]
	\centering
	\caption{Quantitative comparisons:
		removing each specific component from the full PU-GAN pipeline (top five rows) vs. baseline GAN model (6-th row) vs. full PU-GAN pipeline (last row).}
	\label{tab:ablationStudy}
	\resizebox{\linewidth}{!}{
		\begin{tabular}{@{\hspace{1mm}}c@{\hspace{1mm}}||@{\hspace{1mm}}c@{\hspace{3mm}}c@{\hspace{3mm}}c@{\hspace{3mm}}c@{\hspace{3mm}}c@{\hspace{1mm}}|@{\hspace{0.2mm}}c@{\hspace{1mm}}|@{\hspace{1mm}}c@{\hspace{1mm}}|@{\hspace{1mm}}c@{\hspace{1mm}}} \toprule[1pt]
			\multirow{2}*{} & \multicolumn{5}{@{\hspace{1mm}}c@{\hspace{1mm}}|@{\hspace{0.2mm}}}{Uniformity for different $p$ ($10^{-3}$)} & P2F & CD & HD \\ \cline{2-6}
			& 0.4\% & 0.6\% & 0.8\% & 1.0\% & 1.2\% & ($10^{-3}$) & ($10^{-3}$) & ($10^{-3}$)\\ \hline \hline
			Discriminator       &  7.02  & 7.31  & 8.36  &  9.70  & 11.17  & 4.61 & 0.57  & 7.25  \\
			$\mathcal{L_{\text{uni}}}$ & 5.15  & 5.71  & 6.13  & 6.82   & 7.32  & 3.99 & 0.51  & 6.22  \\
			self-attention                &  4.19  & 4.66  & 4.87  & 5.52   & 6.47 & 3.97 & 0.46  & 6.01  \\
			up-down-up             &   3.89 & 4.16 & 4.63 & 5.14 & 5.89 & 3.02 & 0.35	& 5.15\\
			farthest samp.            & 3.56  & 3.76  & 3.78  & 4.15   & 4.97 & 2.72 & 0.31  & 4.96  \\ \hline
			baseline GAN & 8.12  & 8.18  & 8.76  & 9.89   & 11.32  & 4.79 & 0.59  & 7.31   \\ \hline
			full pipeline  & \textbf{3.38} & \textbf{3.49}  & \textbf{3.44}  & \textbf{3.91}  & \textbf{4.64} & \textbf{2.33} & \textbf{0.28}  & \textbf{4.64}  \\ \bottomrule[1pt]
	\end{tabular}}
	\vspace*{-2mm}
\end{table}


\subsection{Ablation Study and Baseline Comparison}
\label{subsec:ablation}

To evaluate the components in PU-GAN, including the GAN framework (\ie, remove the discriminator and keep only the generator), up-down-up expansion unit, self-attention unit, uniform loss, and farthest sampling, we removed each of them from the network and generated upsampling results for the testing models.
Table~\ref{tab:ablationStudy} shows the evaluation results.
Our full pipeline performs the best, and removing any component reduces the overall performance, meaning that each component contributes.
Particularly, removing the GAN framework (\ie, removing the discriminator) causes the largest performance drop, thereby showing the effectiveness of adversarial learning in the framework.

\begin{figure}[!t]
	\centering
	\includegraphics[width=1.0\linewidth]{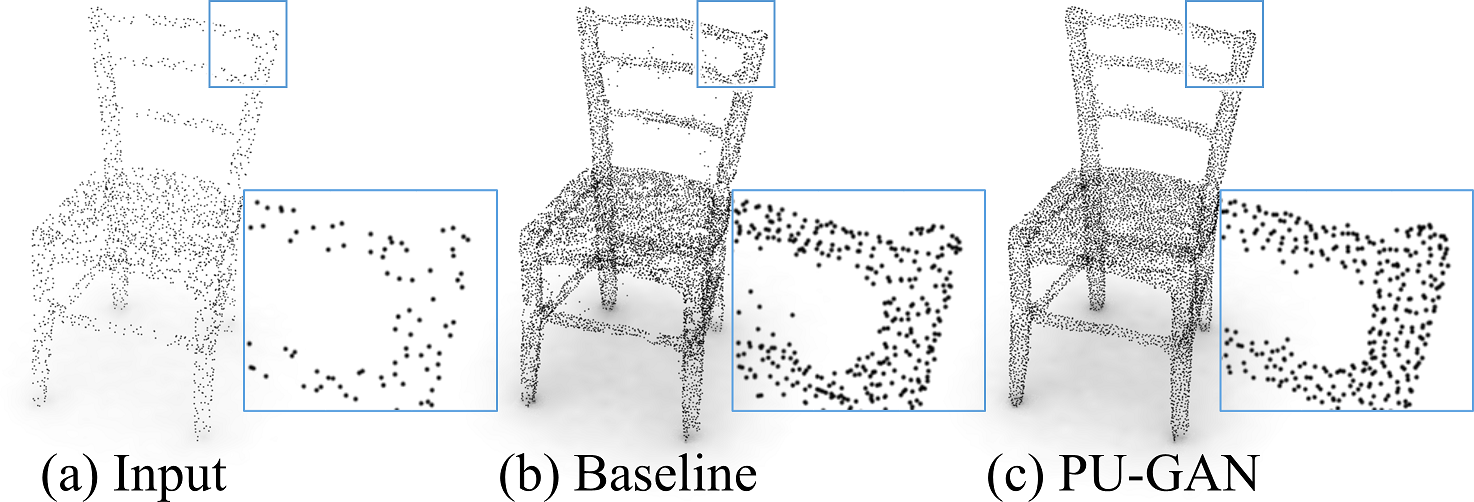}
	\caption{Visual comparison of (c) PU-GAN against (b) the baseline GAN model, given (a) an input of 2,048 non-uniform points.}
	\label{fig:baseline}
	\vspace*{-2mm}
\end{figure}

Further, we designed a GAN baseline for comparison by removing $\mathcal{L_{\text{uni}}}$, self-attention unit, up-down-up expansion unit, and farthest sampling altogether; see supplemental material for its architecture.
Quantitative results shown in the second last row of Table~\ref{tab:ablationStudy} and a visual comparison example shown in Figure~\ref{fig:baseline} demonstrate that simply applying a basic GAN framework is insufficient for the upsampling problem; our adaptations to the GAN model plus the various formulations contribute to realizing a working GAN model.


\subsection{Other Experiments}

\begin{figure}[!t]
	\centering
	\includegraphics[width=1.0\linewidth]{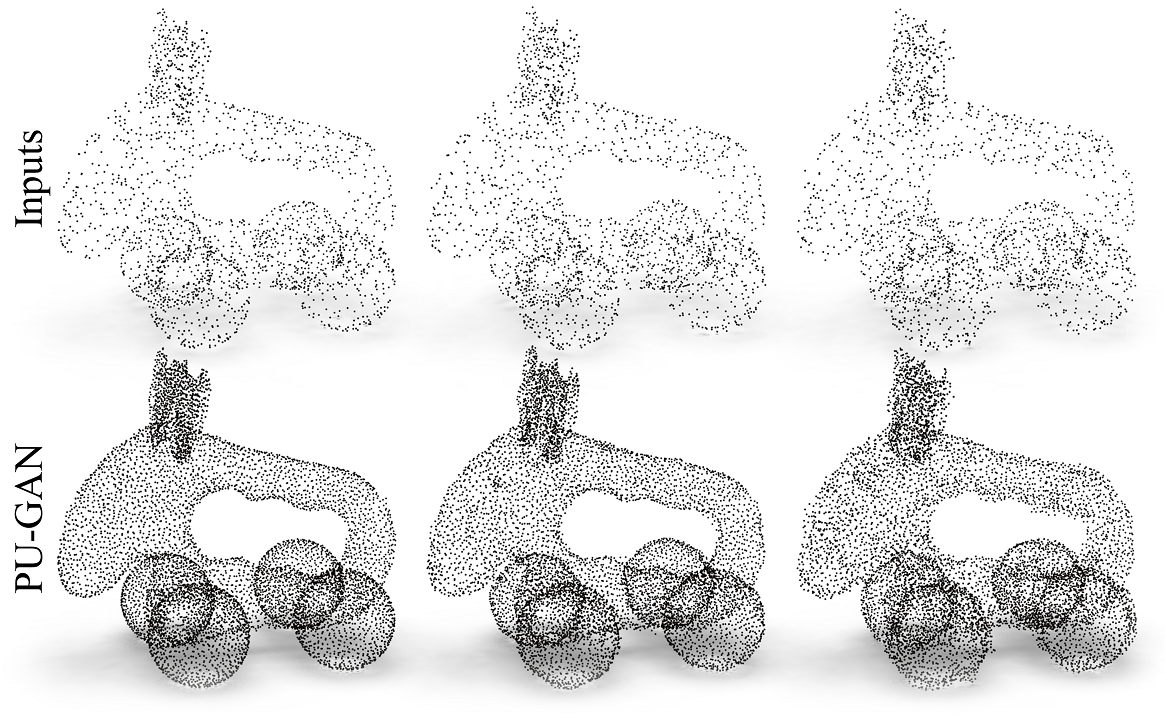}
	\vspace{-4mm}
	\caption{Upsampling results by applying PU-GAN to inputs with noise level of 0, 0.5\%, and 1\%.}
	\label{fig:noise}
\end{figure}

\para{Upsampling point sets of varying noise levels.} \
Figure~\ref{fig:noise} shows the results of using PU-GAN to upsample point sets of increasing Gaussian noise levels, indicating the robustness of PU-GAN to noise and sparsity.
Please refer to the supplemental material for more results.

\begin{figure}[!t]
	\centering
	\includegraphics[width=1.0\linewidth]{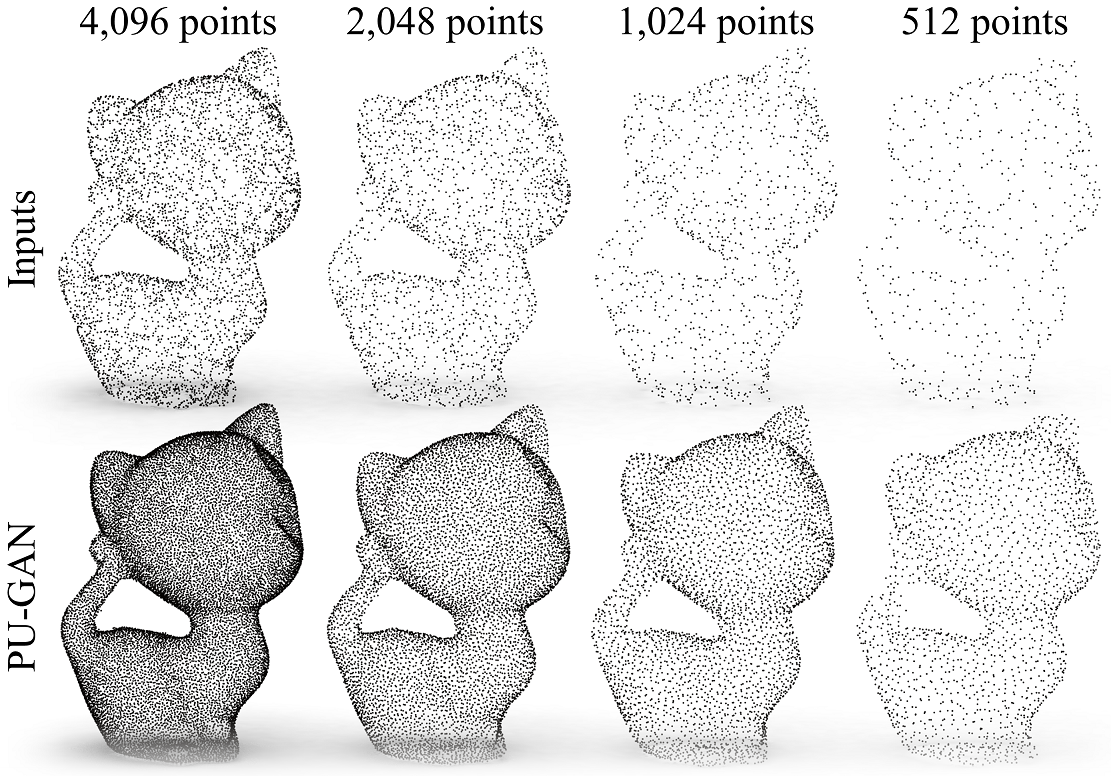}
	\vspace{-4mm}
	\caption{Upsampling point sets (top row) of varying sizes.} 
	\label{fig:diff_point_num}
	\vspace*{-2mm}
\end{figure}

\para{Upsampling point sets of varying sizes.} \
Figure~\ref{fig:diff_point_num} shows the results of upsampling input point sets of different sizes; our method is stable even for input with only 512 points; more results are shown in the supplemental material.

\para{Applications.} \
Besides surface reconstruction and rendering (see Figure~\ref{fig:visualComparison}), point cloud upsampling is also helpful for object recognition.
To demonstrate this, we trained PointNet (vanilla version)~\cite{qi2016pointnet} on ModelNet40 dataset~\cite{wu20153d} for classification under two cases:
in case (i), we directly trained PointNet to take sparse training set (512 points) as inputs for classifying objects in the sparse testing set;
and in case (ii), we upsampled the point clouds in both sparse training and testing sets to 2,048 points using PU-GAN and then trained another PointNet with the upsampled training set for classifying the upsampled data in the testing set.
Results show that, the overall classification accuracy increased from 82.4\% (case (i)) to 83.8\% (case (ii)), indicating that PU-GAN helps improve the classification performance.

\section{Conclusion}
\label{sec:conclusion}

In this paper, we presented, PU-GAN, a novel GAN-based point cloud upsampling network,  that combines upsampling with data amendment capabilities.
Such adversarial network enables the generator to produce a uniformly-distributed point set, and  the discriminator to implicitly penalize outputs that deviate from the expected target.
To facilitate the GAN framework, we introduced an up-down-up unit for feature expansion with error feedback and self-correction, as well as a self-attention unit for better feature fusion.
Further, we designed a compound loss to guide the learning of the generator and discriminator. We demonstrated the effectiveness of our PU-GAN via extensive experiments, showing that it outperforms the state-of-the-art methods for various metrics, and presented the upsampling performance on real-scanned LiDAR inputs.

However, since PU-GAN is designed to complete tiny holes at patch level, it has limited ability in filling large gaps or holes in point clouds; see Figure~\ref{fig:realscan}. Analyzing and synthesizing at the patch level lacks a global view of the overall shape.
\rh{Please refer to the supplemental material for a typical failure case.}
In the future, we will explore a multi-scale training strategy to encourage the network to learn from both local small patched and global structures.
We are also considering exploring conditional GANs, to let the network learn the uniformity and semantic consistency at the same time.

\para{Acknowledgments.} \
We thank anonymous reviewers for the valuable comments.
The work is supported by the 973 Program (Proj. No. 2015CB351706), the National Natural Science Foundation of China with Proj. No. U1613219, the Research Grants Council of the Hong Kong Special Administrative Region (No. CUHK 14203416 \& 14201717), and the Israel Science Foundation grants 2366/16 and 2472/7.

{\small
\bibliographystyle{ieee}
\bibliography{egbib}
}

\end{document}